\documentclass[11pt,a4paper]{article}
\usepackage[hyperref]{acl2017} 
\usepackage{times}
\usepackage{latexsym}
\usepackage{url}
\usepackage{framed}
\usepackage{scrextend}
\usepackage{booktabs}
\usepackage{amsmath}
\usepackage{paralist}
\usepackage{tikz}
\usepackage{color}
\usepackage{tabularx}
\usepackage{multirow}
\usetikzlibrary{positioning}
\usepackage{enumitem}
\usepackage{float}
\usepackage{array,graphicx}
\usepackage{pifont}
\usepackage{diagbox}

\aclfinalcopy 

\title{Turing at SemEval-2017 Task 8: Sequential Approach to Rumour Stance Classification with Branch-LSTM}

\author{
	Elena Kochkina$^1$$^2$, Maria Liakata$^1$$^2$, Isabelle Augenstein$^3$\\
	$^1$ University of Warwick, Coventry, United Kingdom\\
	$^2$ Alan Turing Institute, London, United Kingdom\\
	$^3$ University College London, London, United Kingdom\\
	\{E.Kochkina, M.Liakata\}@warwick.ac.uk, I.Augenstein@ucl.ac.uk
}

\date{}

\begin{document}
\maketitle

\begin{abstract}
This paper describes team Turing's submission to SemEval 2017 RumourEval: Determining rumour veracity and support for rumours (SemEval 2017 Task 8, Subtask A). Subtask A addresses the challenge of rumour stance classification, which involves identifying the attitude of Twitter users towards the truthfulness of the rumour they are discussing. Stance classification is considered to be an important step towards rumour verification, therefore performing well in this task is expected to be useful in debunking false rumours. In this work we classify a set of Twitter posts discussing rumours into either supporting, denying, questioning or commenting on the underlying rumours. We propose a LSTM-based sequential model that, through modelling the conversational structure of tweets, which achieves an accuracy of 0.784 on the RumourEval test set outperforming all other systems in Subtask A. 
\end{abstract}

\section{Introduction}
In stance classification one is concerned with determining the attitude of the author of a text towards a target~\cite{mohammad2016semeval}. Targets can range from abstract ideas, to concrete entities and events. Stance classification is an active research area that has been studied in different domains \cite{ranade2013stance,chuang2015stance}.
Here we focus on stance classification of tweets towards the truthfulness of rumours circulating in Twitter conversations in the context of breaking news. Each conversation is defined by a tweet that initiates the conversation and a set of nested replies to it that form a conversation thread. The goal is to classify each of the tweets in the conversation thread as either \textit{supporting}, \textit{denying}, \textit{querying} or \textit{commenting} (\textit{SDQC}) on the rumour initiated by the source tweet. Being able to detect stance automatically is very useful in the context of events provoking public resonance and associated rumours, as a first step towards verification of early reports \cite{zhao2015enquiring}. For instance, t has been shown that rumours that are later proven to be false tend to spark significantly larger numbers of denying tweets than rumours that are later confirmed to be true \cite{mendoza2010twitter,procter2013readinga,derczynski2014pheme,zubiaga2016analysing}. 

Here we  focus on exploiting the conversational structure of social media threads for stance classification and introduce a novel LSTM-based approach to harness conversations.
\section{Related Work}
\paragraph{Single Tweet Stance Classification}

Stance classification for rumours was pioneered by Qazvinian et al. \shortcite{qazvinian2011rumor} as a binary classification task (support/denial).
Zeng et al. \shortcite{zeng2016unconfirmed} perform stance classification for rumours emerging during crises. Both works use tweets related to the same rumour during training and testing. 

A model based on bidirectional LSTM encoding of tweets conditioned on targets has been shown to achieve state-of-the-art on the SemEval-2016 task 6 dataset \cite{augenstein2016stance}. However the RumourEval task is different as it addresses conversation threads.
\label{sect:data}
\paragraph{Sequential Stance Classification}
 Lukasik et al. \shortcite{lukasik2016hawkes} and Zubiaga et al. \shortcite{zubiaga2016stance} consider the sequential nature of tweet threads in their works. Lukasik et al. \shortcite{lukasik2016hawkes} employ Hawkes processes to classify temporal sequences of tweets. They show the importance of using both the textual content and temporal information about the tweets, disregarding the discourse structure. Zubiaga et al. \shortcite{zubiaga2016stance} model the conversational structure of source tweets and subsequent replies: as a linear chain and as a tree. They use linear- and tree- versions of a CRF classifier, outperforming the approach by Lukasik et al. \shortcite{lukasik2016hawkes}. 

\section{Dataset}
\begin{figure}[t]
	\centering
	\includegraphics[width=\linewidth]{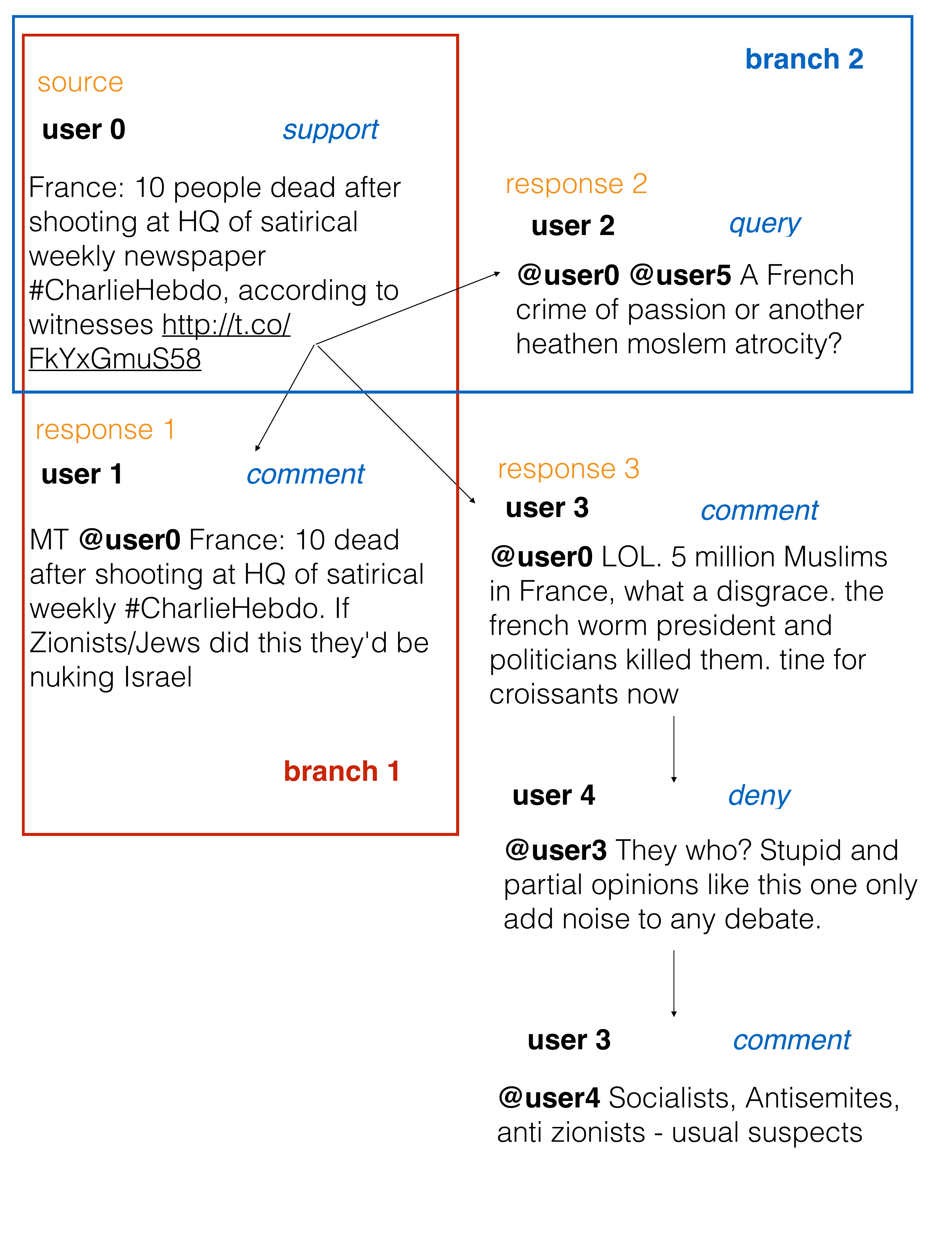}
	\vspace{-1.5cm}
	\caption{Example of a conversation thread from the dataset with three branches, two of which are highlighted. The conversation has a tree structure, which can be split into individual branches by taking each leaf node with all its direct parents.}\vspace{-0.3cm}
	\label{fig:Conversation}
\end{figure}

The dataset provided for this task contains Twitter conversation threads associated with rumours around ten different events in breaking news, including the Paris shootings in Charlie Hebdo, the Ferguson unrest, the crash of a Germanwings plane.  These events include 325 conversation threads consisting of 5568 underlying tweets annotated for stance at the tweet level (breakdown between training, testing and development sets is shown in Table \ref{datastats}) \cite{taskpaper}. 
\begin{table}[th]
\centering
\footnotesize
\begin{tabular}{|l|l|l|l|}
\hline
            & \textbf{\# threads} &  \textbf{\# branches} &  \textbf{\# tweets}     \\ \hline
 Development & 25         & 215         & 281        \\ \hline
 Testing     & 28         & 772         & 1049      \\ \hline
 Training    & 272        & 3030        & 4238      \\ \hline
  \textbf{Total}       & 325        & 4017        & 5568       \\ \hline
\end{tabular}%
\caption{Number of threads, branches and tweets in the training, development and testing sets.}
\label{datastats}
\end{table}

\begin{table}[th]
\centering
\footnotesize
\begin{tabular}{|l|l|l|l|l|}
\hline
             &  \textbf{S}    &  \textbf{D}   &  \textbf{Q}   &  \textbf{C}    \\ \hline
 Development       & 69   & 11  & 28  & 173  \\ \hline
 Testing         & 94   & 71  & 106 & 778  \\ \hline
Training         & 841  & 333 & 330 & 2734 \\ \hline
 \textbf{Total}             & 1004 & 415 & 464 & 3685 \\ \hline
\end{tabular}%
\caption{Per-class distribution of tweets in the training, development and testing sets.} \vspace{-0.2cm}
\label{class}
\end{table}

Each thread includes a source tweet that initiates a conversation and nested tweets responding to either the source tweet or earlier replies.  The thread can be split into linear branches of tweets, where a branch is defined as a chain of tweets that starts with a leaf tweet including its direct parent tweets, all the way up to the source tweet. Figure \ref{fig:Conversation} shows an example of a conversation along with its annotations represented as a tree structure with highlighted branches. The depth of a tweet is the number of its parents starting from the root node. Branches 1 and 2 in Figure \ref{fig:Conversation} have depth one whereas branch 3 has depth three.
There is a clear class imbalance in favour of \textit{commenting}  tweets (66\%) and \textit{supporting} tweets (18\%), whereas the \textit{denying} (8\%) and \textit{querying} classes (8\%) are under-represented (see Table \ref{class}). While this imbalance poses a challenge, it is also indicative of the realistic scenario where only a few users question the veracity of a statement. 
 
\section{System Description}
\subsection{Features}
\label{sect:feat} 
Prior to generating features for the tweets, we perform a pre-processing step where we remove non-alphabetic characters, convert all words to lower case and tokenise texts.\footnote{For implementation of all pre-processing routines we use Python 2.7 with the NLTK package.} Once tweet texts are pre-processed, we extract the following features:
\begin{figure}
  \begin{framed}
  
      \textbf{[As querying]} @username Weren't you the one who abused her? \vspace{0.15cm}
      
      \textbf{[As supporting]} "Go online \&amp; put down 'Hillary Clinton illness,'" Rudy says. Yes -- but look up the truth -- not health smears  https://t.co/EprqiZhAxM \vspace{0.15cm}
      
     \textbf{[As supporting]} @username I demand you retract the lie that people in \#Ferguson were shouting "kill the police", local reporting has refuted your ugly racism \vspace{0.15cm}
     
     \textbf{[As commenting]} @FoxNews six years ago... real good evidence.  Not! 
  
  \end{framed}
  \caption{Examples of misclassified denying tweets. } \vspace{-0.2cm}
  \label{fig:deny}
 \end{figure} 
\begin{compactitem}
    \item \textbf{Word vectors:} we use a word2vec \cite{mikolov2013efficient} model pre-trained on the Google News dataset (300d)  \footnote{We have also tried using Glove word embeddings trained on Twitter dataset, but it lead to a decrease in performance on both development and testing sets comparing to the Google News word vectors} using the gensim package \cite{rehurek_lrec}. 
	\item \textbf{Tweet lexicon:} (1) count of negation words\footnote{A presence of any of the following words would be considered as a presence of negation: not, no, nobody, nothing, none, never, neither, nor, nowhere, hardly, scarcely, barely, don't, isn't, wasn't, shouldn't, wouldn't, couldn't, doesn't} and (2) count of swear words.\footnote{A list of 458 bad words was taken from http://urbanoalvarez.es/blog/2008/04/04/bad-words-list/} 
	\item \textbf{Punctuation:} (1) presence of a period, (2) presence of an exclamation mark, (3) presence of a question mark, (4) ratio of capital letters.
	\item \textbf{Attachments:} (1) presence of a URL and (2) presence of images.
	\item \textbf{Relation to other tweets} (1) Word2Vec cosine similarity wrt source tweet, (2) Word2Vec cosine similarity wrt preceding tweet, and (3) Word2Vec cosine similarity wrt thread
	\item \textbf{Content length:} (1) word count and (2) character count.
	\item \textbf{Tweet role:} whether the tweet is a source tweet of a conversation.
\end{compactitem}
Tweet representations are obtained by averaging word vectors in a tweet and then concatenating with the additional features into a single vector, at the preprocessing step. This set of features have shown to be the best  comparing to using word2vec features on their own or any of the reduced combinations of these features.
\begin{table*}[t]
 	\centering
 	\begin{tabular}{ | l | l | l | l | l | l | l | }
 		\hline
	 &  \textbf{Accuracy} &  \textbf{Macro F} &  \textbf{S} &  \textbf{D} &  \textbf{Q} &  \textbf{C} \\ \hline
	Development & 0.782 & 0.561 & 0.621 & 0.000 & 0.762 & 0.860 \\ \hline
	Testing &  \textbf{0.784} & 0.434 & 0.403 & 0.000 & 0.462 & 0.873 \\ \hline
 	\end{tabular}
 	\caption{Results on the development and testing sets. Accuracy and F1 scores: macro-averaged and per class (S: \textit{supporting}, D: \textit{denying}, Q: \textit{querying}, C: \textit{commenting}).}
 	\label{result_main}
 \end{table*}
 \begin{table*}[]
 	\centering
 	\footnotesize
 	\resizebox{\textwidth}{!}{%
 	\begin{tabular}{ | l | l | l | l | l | l | l | l | l | l | l | l | }
\hline
	\textbf{Depth} & \textbf{\# tweets} & \textbf{\# S} & \textbf{\# D} & \textbf{\# Q} & \textbf{\# C} & \textbf{Accuracy} & \textbf{MacroF} & \textbf{S} & \textbf{D} & \textbf{Q} & \textbf{C} \\ \hline
	0 & 28 & 26 & 2 & 0 & 0 & 0.929 & 0.481 &  \textbf{0.963} & 0.000 & 0.000 & 0.000 \\ \hline
	1 & 704 & 61 & 60 & 81 & 502 & 0.739 & 0.348 & 0.000 & 0.000 &  \textbf{0.550} &  \textbf{0.842} \\ \hline
	2 & 128 & 3 & 6 & 7 & 112 & 0.875 & 0.233 & 0.000 & 0.000 & 0.000 &  \textbf{0.933} \\ \hline
	3 & 60 & 2 & 1 & 5 & 52 & 0.867 & 0.232 & 0.000 & 0.000 & 0.000 &  \textbf{0.929} \\ \hline
	4 & 41 & 0 & 0 & 3 & 38 & 0.927 & 0.481 & 0.000 & 0.000 & 0.000 &  \textbf{0.962} \\ \hline
	5 & 27 & 1 & 0 & 1 & 25 & 0.926 & 0.321 & 0.000 & 0.000 & 0.000 &  \textbf{0.961} \\ \hline
	6+ & 61 & 1 & 2 & 9 & 49 & 0.803 & 0.223 & 0.000 & 0.000 & 0.000 &  \textbf{0.891}\\ \hline
\end{tabular}}
 	\caption{Number of tweets per depth and performance at each of the depths.} 
 	\label{result_depth}
 \end{table*}
\subsection{Branch - LSTM Model}
\begin{figure}
	\centering
	\includegraphics[width=\linewidth]{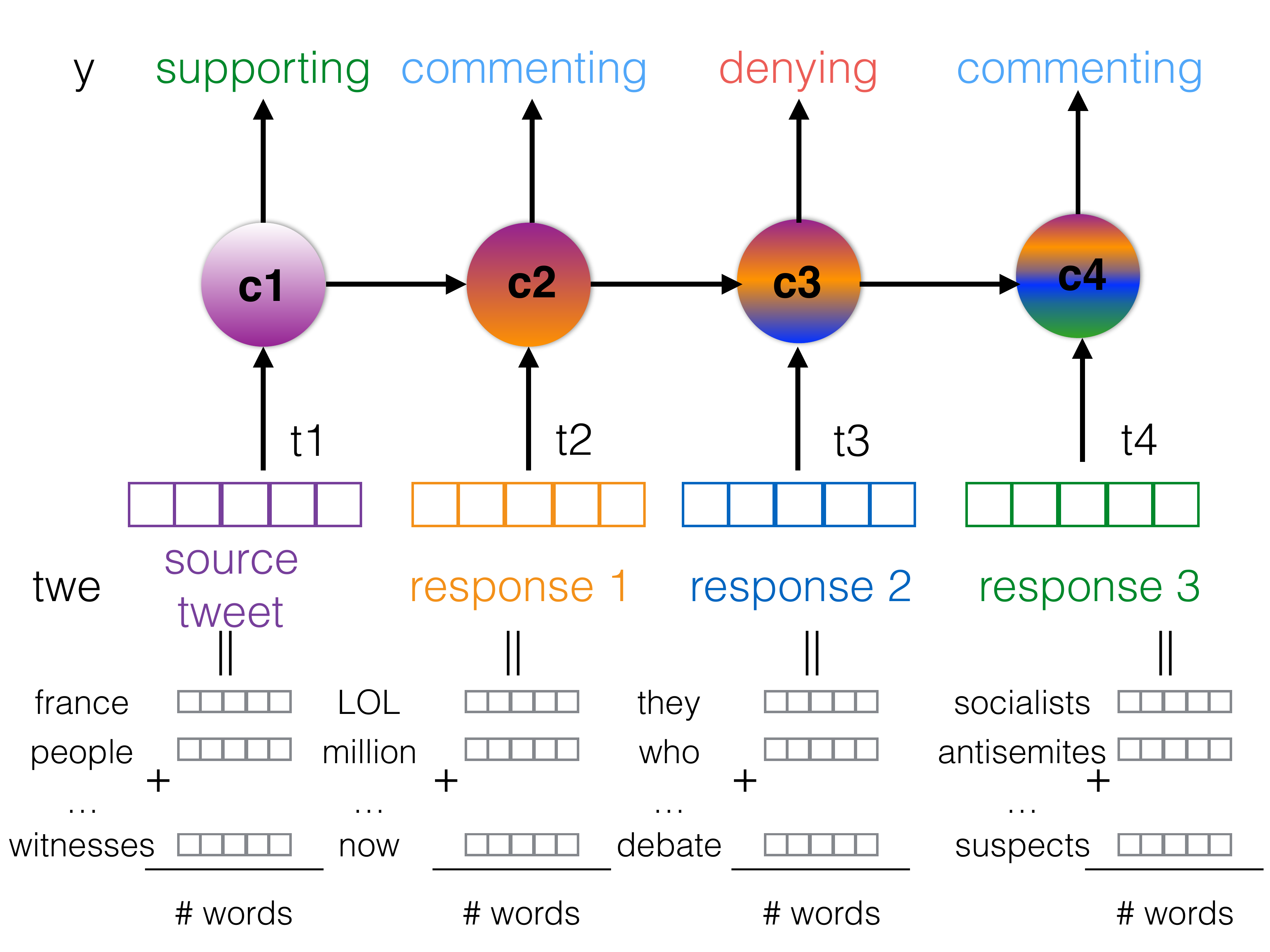}
	\vspace{-0.9cm}
	\caption{Illustration of the input/output structure of the branch-nestedLSTM model.} \vspace{-0.2cm}
	\label{fig:hlstm}
\end{figure}

 To tackle the task of rumour stance classificaiton, we propose \textit{branch-LSTM}, a neural network architecture that uses layers of LSTM units \cite{hochreiter1997long} to process the whole branch of tweets, thus incorporating structural information of the conversation (see the illustration of the \textit{branch-LSTM} on the Figure \ref{fig:hlstm}). The input at each time step $i$  of the LSTM layer is the representation of the tweet as a vector.  We record the output of each time step so as to attach a label to each tweet in a branch\footnote{For implementation of all models we used Python libraries Theano \cite{bastien2012theano} and Lasagne \cite{lasagne}.}. This output is fed through several dense ReLU layers, a 50\% dropout layer,  and then through a softmax layer to obtain class probabilities. 
We use zero-padding and masks to account for the varying lengths of tweet branches. The model is trained using the categorical cross entropy loss function.
Since there is overlap between branches originating from the same source tweet, we exclude the repeating tweets from the loss function using a mask at the training stage.
The model uses tweet representation as the mean average of word vectors concatenated with extra features described above. Due to the short length of tweets, using more complex models for learning tweet representations, such as an LSTM that takes each word as input at each time step and returns the representation at the final time step, does not lead to a noticeable difference in the performance based on cross-validation experiments on the training and development sets, while taking significantly longer to train.
We experimented with replacing the unidirectional LSTMs with bidirectional LSTMs but could observe no improvements in accuracy (using cross-validation results on the training and development set). 
\section{Experimental Setup}
The dataset is split into training, development and test sets by the task organisers.
 We determined the optimal set of hyperparameters via testing the performance of our model on the development set for different parameter combinations. We used the Tree of Parzen Estimators (TPE) algorithm \footnote{We used the implementation of the TPE algorithm in the hyperopt package \cite{bergstra2013making}} to search the parameter space, which is defined as follows: the number of dense ReLU layers varies from one to four; the number of LSTM layers is one or two; the mini-batch size is either 32 or 64; the number of units in the ReLU layer is one of \{100, 200, 300, 400, 500\}, and in the LSTM layer one of \{100, 200, 300\}; the strength of the L2 regularisation is one of \{0.0, 1e-4, 3e-4, 1e-3\} and the number of epochs is selected from \{30, 50, 70, 100\}. We performed 100 trials of different parameter combinations optimising for accuracy on the development set in order to choose the best combination.  We fixed hyperparameters to train the model on combined training and development sets and evaluated on the held out test set.

 \section{Results}
\label{resultssection}
The performance of our model on the testing and development set is shown in Table \ref{result_main}. 
Together with the accuracy we show macro-averaged F-score and per-class macro-averaged F-scores as these metrics account for the class imbalance. The difference in accuracy between testing and development set is minimal, however we see significant difference in Macro-F score due to different class balance in these sets. Macro-F score could be improved if we used it as a metric for optimising hyper-parameters. The \textit{branch-LSTM} model predicts \textit{commenting}, the majority class well, however it is unable to pick out any \textit{denying}, the most-challenging under-represented class. Most \textit{denying} instances get misclassified as \textit{commenting} (see Table \ref{confusion}), with only one tweet misclassified as \textit{querying} and two as \textit{supporting} (Figure \ref{fig:deny}). An increased amount of labelled data would be helpful to improve performance of this model.  
\begin{table}[]
 	\centering
 	\footnotesize
 	\begin{tabular}{ | l | l | l | l | l | }
\hline
	 \backslashbox{\textbf{Label}}{\textbf{Prediction}} &  \textbf{C} &  \textbf{D} &  \textbf{Q} &  \textbf{S} \\ \hline
	 \textbf{Commenting} & 760 & 0 & 12 & 6 \\ \hline
	 \textbf{Denying} & 68 & 0 & 1 & 2 \\ \hline
	 \textbf{Querying} & 69 & 0 & 36 & 1 \\ \hline
	 \textbf{Supporting} & 67 & 0 & 1 & 26 \\ \hline
 	\end{tabular}
 	\caption{Confusion matrix for testing set predictions} \vspace{-0.2cm}
 	\label{confusion}
 \end{table}
As we were considering conversation branches, it is interesting to analyse the performance distribution across different tweet depths (see Table \ref{result_depth}). Maximum depth/branch length in the testing set is 13 with most tweets concentrated at depths from 0 to 3. Source tweets (depth zero) are usually \textit{supporting} and the model predicts these very well, but performance of \textit{supporting} tweets at other depths decreases.  
The model does not show a noticeable difference in performance on tweets of varying lengths. 
 
\section{Conclusions}
This paper describes the Turing system entered in the SemEval-2017 Task 8 Subtask A. Our method decomposes the tree structure of conversations into linear sequences and achieves accuracy 0.784 on the testing set and sets the state-of-the-art for rumour stance classification. In future work we plan to explore different methods for modelling tree-structured conversations. 
\section*{Acknowledgments}
This work was supported by The Alan Turing Institute under the EPSRC grant EP/N510129/1. Cloud computing resource were kindly provided through a Microsoft Azure for Research Award. Work by Elena Kochkina was partially supported by the Leverhulme Trust through the Bridges Programme.

\bibliography{semeval2017}
\bibliographystyle{acl_natbib}

\end{document}